\documentclass[letterpaper]{article} % DO NOT CHANGE THIS
\usepackage{aaai23}
\usepackage{times}  % DO NOT CHANGE THIS
\usepackage{helvet}  % DO NOT CHANGE THIS
\usepackage{courier}  % DO NOT CHANGE THIS
\usepackage[hyphens]{url}  % DO NOT CHANGE THIS
\usepackage{graphicx} % DO NOT CHANGE THIS
\usepackage{natbib}  % DO NOT CHANGE THIS AND DO NOT ADD ANY OPTIONS TO IT
\usepackage{caption} % DO NOT CHANGE THIS AND DO NOT ADD ANY OPTIONS TO IT
\usepackage{algorithm}
\usepackage{algorithmic}
\usepackage{mathtools}
\usepackage{amsmath}
\usepackage{amsthm}
\usepackage{amsfonts}
\usepackage{subfigure}
\usepackage{multirow}
\usepackage{siunitx}

\usepackage{newfloat}
\usepackage{listings}
\DeclareCaptionStyle{ruled}{labelfont=normalfont,labelsep=colon,strut=off} % DO NOT CHANGE THIS
\floatstyle{ruled}
\newfloat{listing}{tb}{lst}{}
\floatname{listing}{Listing}
\title{Multi-modal Transformer Path Prediction for Autonomous Vehicle}
\author{
Chia Hong Tseng\textsuperscript{\rm 1}, 
Jie Zhang\textsuperscript{\rm 1}, 
Min-Te Sun\textsuperscript{\rm 1}, 
Kazuya Sakai\textsuperscript{\rm 2}, 
Wei-Shinn Ku\textsuperscript{\rm 3}
}
\affiliations{
    \textsuperscript{\rm 1} National Central University, Taiwan\\
    \textsuperscript{\rm 2} Tokyo Metropolitan University, Japan\\
    \textsuperscript{\rm 3} Auburn University, United States of America
    
}

\usepackage{bibentry}
\UseRawInputEncoding
\begin{document}

\maketitle

\begin{abstract}
Reasoning about vehicle path prediction is an essential and challenging problem for the safe operation of autonomous driving systems. There exist many research works for path prediction. However, most of them do not use lane information and are not based on the Transformer architecture. By utilizing different types of data collected from sensors equipped on the self-driving vehicles, we propose a path prediction system named Multi-modal Transformer Path Prediction (MTPP) that aims to predict long-term future trajectory of target agents. To achieve more accurate path prediction, the Transformer architecture is adopted in our model. To better utilize the lane information, the lanes which are in opposite direction to target agent are not likely to be taken by the target agent and are consequently filtered out. In addition, consecutive lane chunks are combined to ensure the lane input to be long enough for path prediction. An extensive evaluation is conducted to show the efficacy of the proposed system using nuScene, a real-world trajectory forecasting dataset.
\end{abstract}

\section{Introduction}
In recent years, many works have been proposed to focus in the area of autonomous driving. Among them, path prediction is one of the important topics. There are many existing methods for target agent path prediction, including rule-based, Bayesian, and learning-based models. Unfortunately, most of these works use map input instead of lane input as their feature. Compared with lane input, map input does not provide the direction information, which may lead to inaccurate prediction results. Moreover, most of the models in these existing works are based on LSTM~\cite{LSTM} instead of Transformer~\cite{vaswani2017attention}. Transformer is better than LSTM for two reasons. First, Transformer is able to extract high quality features by taking the whole sequence of data into consideration. Second, Transformer does not need a looping structure. As a result, Transformer is more appropriate for auto-regression method. In self-driving applications, the inference time is crucial because it allows the vehicle to react more quickly to dangerous events.

In our research, we intend to design a system which can accurately predict the future positions of a target agent from its past history. There are several challenges in this research. First, the error of the path prediction increases quickly as the prediction horizon increases. It is hard to design a model to achieve competitive prediction results on the nuScene~\cite{nuScenes} leaderboard. Second, although the nuScene dataset provides the lane information, it misses some critical information (e.g. missing labels). It is difficult to properly prepare the lane input. Finally, due to the nature of the spatial database, the lane information always comes in small chunks. A single lane is usually too short for a path prediction of 6 seconds. It requires extra effort to put several consecutive lanes together for lane input. 

In this thesis, to address these challenges, we design a path prediction system named Multi-modal Transformer Path Prediction (MTPP). To overcome the first challenge, we adopt the Transformer architecture in the model, which can effectively encode and decode long sequence data. In addition, we add lane information to model input, which can confine the prediction result in a more reasonable area in the last few seconds. For the second challenge, we filter out the lanes which are in the opposite direction to target agent and apply lane mask to remove the missing lane from consideration in path prediction. To deal with the third challenge, we carefully stitch the consecutive lane chunks together for lane input. In summary, the contributions of this research are listed as follows.

\begin{itemize}
    \item 
    We build a system, namely MTPP, which predicts three paths for each target agent. Among them, the most likely path is chosen as the future trajectory for the target agent.
    \item
    To produce more accurate long-term path prediction, Transformer architecture is used in our model to effectively encode and decode long sequence data.
    \item
    Some lanes may have different directions from the target agent. We filter out these lanes and apply lane mask to remove the missing lane from consideration in path prediction.
    \item
    Each lane chunk in nuScene can be as short as 20 meters, which is not long enough for path prediction of 6 seconds. We carefully combine consecutive lane chunks together to create the lane input with an appropriate length.
    \item
    We conduct extensive experiments to illustrate that our MPTT system is better than Trajectron++~\cite{Trajectron++} and the lane input helps to achieve more accurate path prediction.
\end{itemize}

\section{Related Work}
Motion prediction task has been one of the hot research topics in recent years, thanks to the popularity of self-driving cars~\cite{waymo} and those public autonomous driving datasets~\cite{nuScenes,Argoverse,waymo}. In this chapter, we break down the task of motion prediction into two different aspects: Model Structure and Map Representation. They are elaborated in detail in the following subsections.

\subsection{Model Structure}
The works in vehicle motion prediction can be categorized into rule-based models, Bayesian models, and learning-based models. Some earlier works in vehicle motion prediction are rule-based models. In~\cite{6696982}, Constant Yaw Rate and Acceleration (CYRA) and Maneuver Recognition Module (MRM) are applied to obtain the future trajectories of vehicles, In~\cite{Fuzzy_Control}, a set of self-defined rules are used to predict lane change trajectory.

Other earlier works in vehicle motion prediction are Bayesian-based models. In~\cite{4282823}, Kalman filter is used to predict the next position or velocity information of vehicles. In~\cite{hmm}, Hidden Markov Model (HMM) is used to first predict multiple possible trajectories of a vehicle, then the cognitive distance space is used to choose the most likely trajectory. ~\cite{based_on_driver_behavior_estimation} design lane-change behavior classification using HMM and analysis based on real driving dataset to generate vehicle trajectory. These rule-based and Bayesian models may perform well in short-term trajectory prediction, but not in long-term trajectory prediction because the driver maneuvers may be influenced by many complex factors.

Many recent works in vehicle motion prediction turn to use learning models. Some of them use only target histories as model input to predict the future trajectories~\cite{8317913,8190764}. In these approaches, the models are able to capture the relationship between histories and future trajectories. However, those works ignore other types of information, such as interactions with surrounding environments. Some other works use heterogeneous inputs such as map information, neighbor agents, and turn signal. In~\cite{Deo_2018_CVPR_Workshops}, Deo and Trivedi map the surroundings on a grid and use a Convolution Neural Network (CNN) to predict the agent trajectory. ~\cite{Trajectron++} use heterogeneous inputs for Conditional Variational AutoEncoder (CVAE) to predict trajectories.  ~\cite{map-prediction} use heterogeneous inputs for a Graph Neural Network (GNN) to predict goal points and generate trajectories for goal points.

\subsection{Map Representation}
Another key design choice is map representation and how it is encoded. We categorize these map representations into two categories, i.e., rasterized top-down map, and vectorized map. In~\cite{nuScenes,Argoverse,waymo}, the annotations of HD maps are in the form of occupancy grid map with multiple channels (e.g. lanes, intersections, drivable areas, sidewalks, etc.). The works in~\cite{Trajectron++,Deo_2018_CVPR_Workshops} use a multi-channel rasterized top-down map as CNN input to capture the road graph environment. However, this approach sacrifices the direction information in lanes, which will lead to non-compliant predictions. Other works~\cite{map-prediction,lapred} extract the center line information and represent lanes as vectorized input for deep learning model. This approach learns the road graph environment and outputs a number of trajectories up to the number of possible future lanes.

\section{Preliminary}
To do vehicle trajectory predictions we need to go through objects detection, objects tracking, and vehicle trajectory prediction. In this chapter, we will introduce methods we use in each of the following sections.

\subsection{Dataset}

\subsubsection{nuScenes}
The nuScenes dataset is a public large-scale dataset for self-driving application development. This dataset includes 1000 scenes collected in urban environment in Boston and Singapore. Each scene lasts 20 seconds and contains data collected by 13 sensors such as 6 cameras (3 forward direction and 3 backward direction, all in 16Hz sample rate), 5 RADAR sensors (13Hz sample rate), LiDAR (on top of the ego vehicle, 20Hz sample rate), and inertial measurement unit (IMU). The annotations of nuScenes comes in 2Hz over the entire dataset. nuScenes has been used for lots of competitions in the research of autonomous driving, such as 3D object Detection~\cite{yolov3}, 3D tracking~\cite{tracking}, and Trajectory prediction~\cite{Trajectron++}. nuScenes is the first public large-scale dataset that provides $360^\circ$ sensor coverage for data collection and the first dataset to include nighttime and rainy conditions. 

The nuScene dataset also provides their toolkit for users to utilize their dataset, making it easier for everyone to train their own model.

\section{Proposed Method}

\begin{figure}[!ht]
    \centering
    \includegraphics[width=0.4\textwidth]{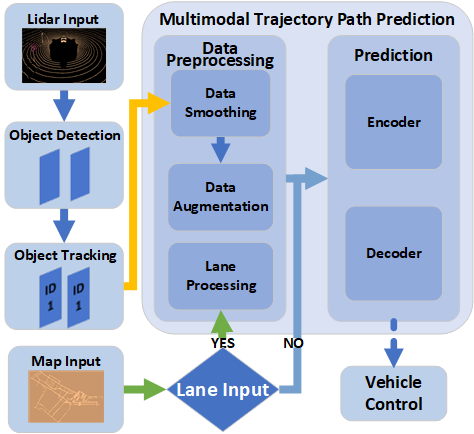}
    \caption{Multi-modal Transformer Path Prediction Architecture} 
    \label{fig:MultiModal Transformer Architecture}
\end{figure}

Figure~\ref{fig:MultiModal Transformer Architecture} illustrates the architecture of Multi-modal Transformer Path Prediction (MTPP) system. Depending on the availability of lane information in the map, our system has two modes. When the map has only the drivable area, our system is able to avoid predicting cars in the non-drivable area. When the map further contains lane information, our system takes advantage of that information so that the cars are predicted to follow the lane most of the time. As shown in Figure~\ref{fig:MultiModal Transformer Architecture}, the system requires LiDAR and Map inputs. The LiDAR input is passed to Object Detection and Object Tracking, and the result will be processed by Data Smoothing and Data Augmentation. If the Map input contains lane information, the Map input will be passed to Lane Processing. The processed LiDAR and Map data will be fed to Prediction model. Finally, the predicted trajectories will be sent to Vehicle Control module to enhance the driving safety.

\subsection{Data pre-processing}
In data pre-processing, three steps are performed to obtain the training data, which include Data Smoothing, Data Augmentation, and Lane Processing. The following sections describe these steps in detail.
    \subsubsection{Data Smoothing}
    The labeled data may still contain some noise. To fix this issue, the Kalman filter~\cite{kalman-filter} is used to smooth nuScene data. This data smoothing helps the model focus on learning the features of the data instead of the noise.
    \subsubsection{Data Augmentation}
    In deep learning, it is important to collect a large labeled dataset for model training. However, in reality, it is hard to collect a large labeled dataset. One common practice to enlarge the training dataset is data augmentation. In this research, two methods of augmentation are used, rotation and upsampling.
    The scene is rotated from $0^\circ$ to $360^\circ$ with the increment of $15^\circ$ per step. By doing so, the dataset is enlarged by 24 times.
    The nuScene dataset is an unbalanced dataset. The number of cars going straight is roughly six times of the number of cars taking turns.
    To deal with unbalanced dataset, the cars taking turns are upsampled six times. 

    \subsubsection{Lane Processing}
    In this module, the lane information is extracted by nuScense devkit~\cite{nuScenes} and some cars and lanes are filtered.
    If a target agent is not located in the lane region (e.g., parked cars), the target agent will be removed from the dataset. In addition, the following three steps are applied so that at most three lanes are taken into consideration in the trajectory prediction, including direction filtering, nearest three-lane identification, and lane extension. These steps are elaborated as follows.
    
\begin{figure}[!ht]
    \begin{center}
    \includegraphics[width=0.35\textwidth]{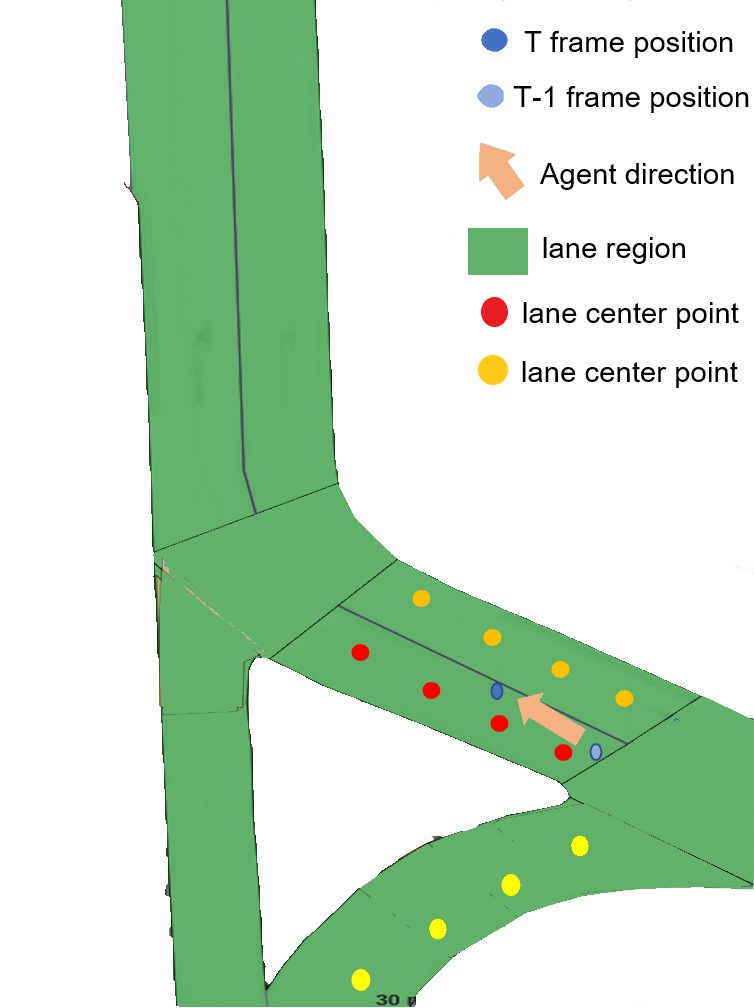}
    \caption{Lane Processing} 
    \label{fig:lane_process_1}
    \end{center}
\end{figure}
    
\begin{figure}[!ht]
    \begin{center}
    \includegraphics[width=0.35\textwidth]{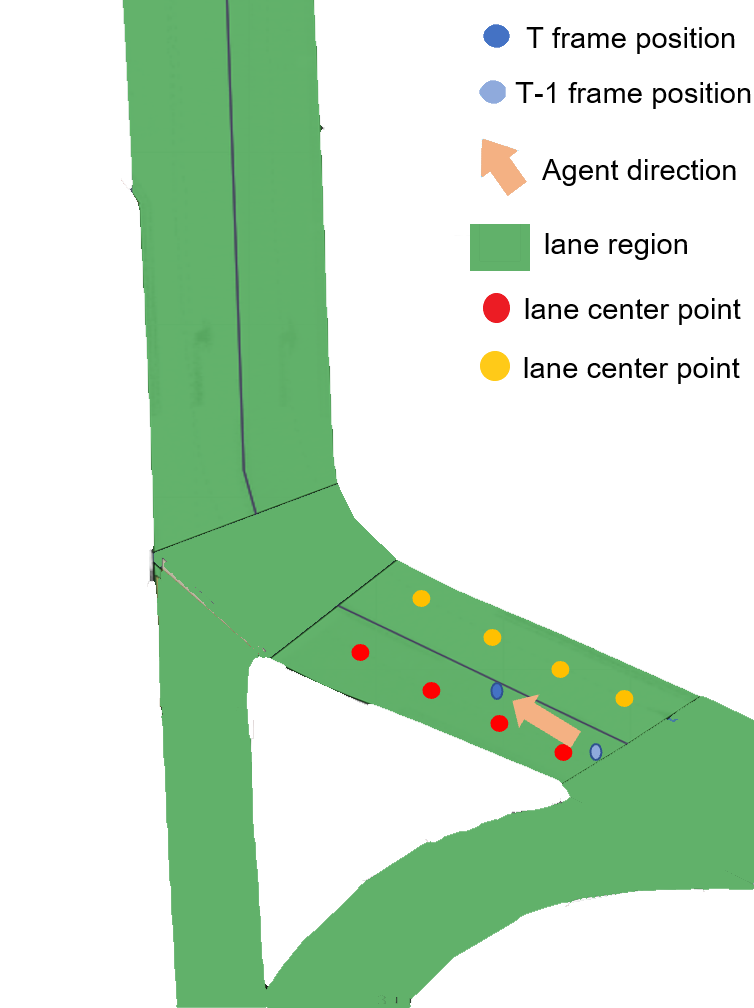}
    \caption{Lane Processing} 
    \label{fig:lane_process_2}
    \end{center}
\end{figure}
    
\begin{figure}[!ht]
    \begin{center}
    \includegraphics[width=0.35\textwidth]{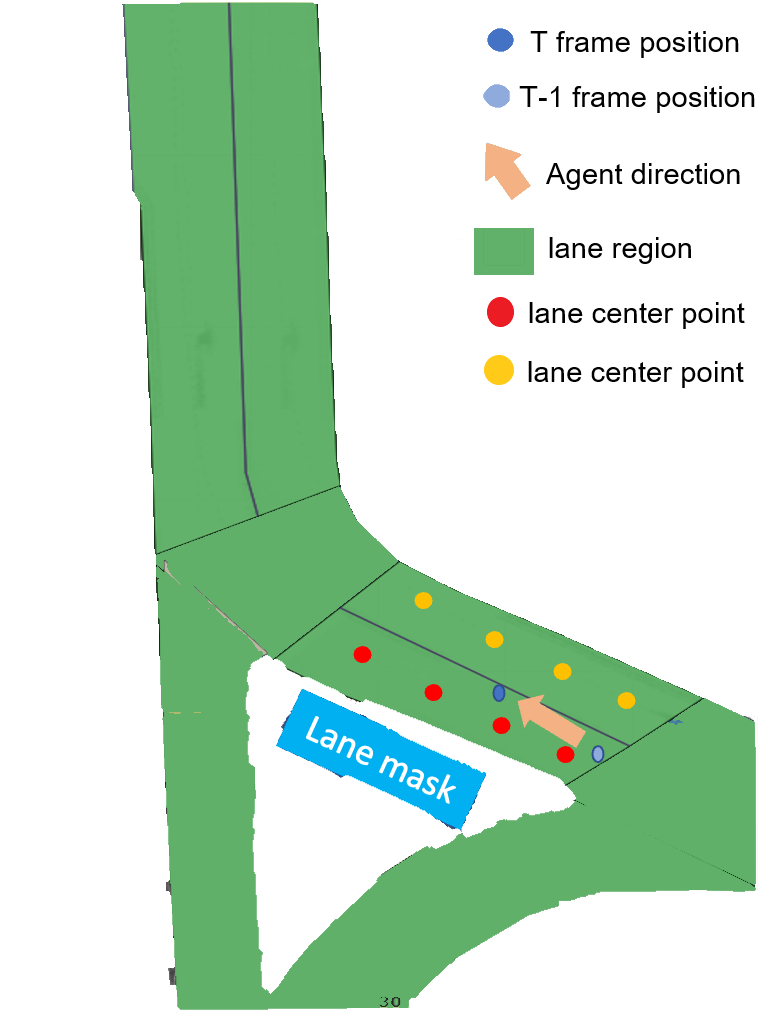}
    \caption{Lane Processing} 
    \label{fig:lane_process_3}
    \end{center}
\end{figure}
    
\begin{figure}[!ht]
    \begin{center}
    \includegraphics[width=0.35\textwidth]{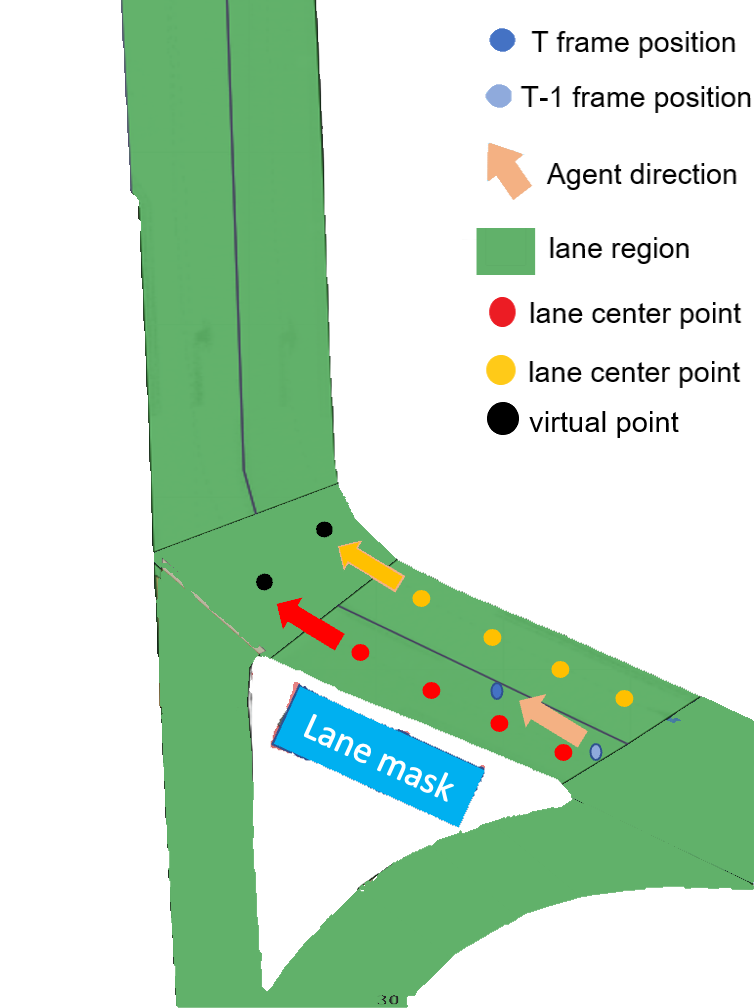}
    \caption{Lane Processing} 
    \label{fig:lane_process_4_extend}
    \end{center}
\end{figure}
        
\begin{figure}[!ht]
    \begin{center}
    \includegraphics[width=0.35\textwidth]{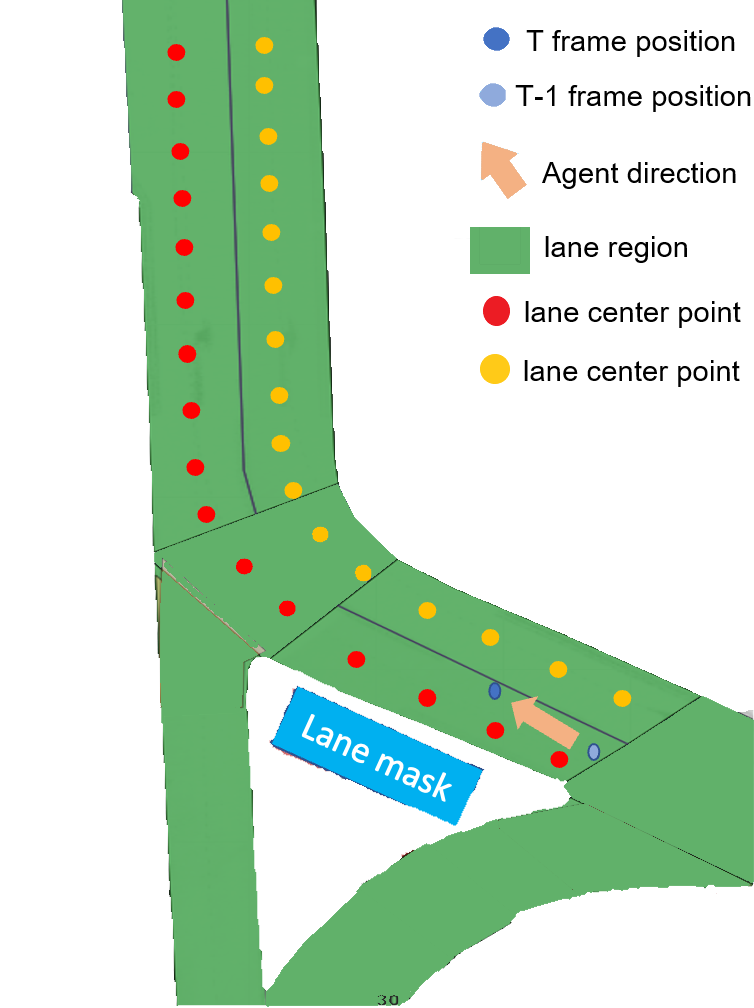}
    \caption{Lane Processing} 
    \label{fig:lane_process_4_get_extend}
    \end{center}
\end{figure}
    
\begin{enumerate}
    \item \textbf{Direction filtering} - Using the current coordinates of a target agent, all the surrounding lanes can be obtained by nuScene devkit. However, some of these lanes may not be suitable for the target agent because they are not in the same direction of the target agent. 
        
    With Figure~\ref{fig:lane_process_1} as an example, the historical coordinates of the target agent (marked by blue) are used to calculate the direction of the target agent. To be exact, the  direction of a target agent in $t$-th frame, denoted by $\vec{X_t}$, is the difference of the coordinates of the target agent in $t$-th frame and $t-1$-th frame. On the contrary, the information of a lane, obtained by nuScene devkit, contains a set of lane center points $L = \{l_1,l_2,\dots,l_{n-1},l_n\}$,  where $l_i$ denotes the $i$-th center point. The distance between adjacent center points is 5 meters. The lane direction of $l_i$, denoted as $\vec{L_i}$, is calculated by Equation~\ref{eq2}.
    
    \begin{equation}\label{eq2}
    \centering
    \begin{split}
    \Vec{L_i} &=
    \begin{cases}
    l_i - l_{i-1} & \mathrm{if}\ 2 \leq i \leq n,\\
    \Vec{L_2} & \mathrm{if}\ i = 1
    \end{cases}
    \end{split}
    \end{equation}
    
    Given a target agent and a lane in $t$-th frame, the direction of the lane for the target agent is $\vec{L_j}$, where $l_j$ is the nearest center point to the target agent. If the angle spanned between $\vec{X_t}$ and $\vec{L_j}$ is larger than $30^\circ$, the lane is considered as in the wrong direction and is subsequently removed from the consideration of trajectory prediction of the target agent. An example is shown in Figure~\ref{fig:lane_process_2}, where the lane marked in yellow is removed because the angle spanned between the directions of the target agent and the lane is larger than $30^\circ$.
        
    \item \textbf{Nearest three-lane identification} - After the lanes in the wrong direction are removed, there may still be quite a few lanes for consideration. In general, the input size of a deep neural network model is fixed. As a consequence, the number of the lanes for the input of our model is set to be 3. First, the lane closest to target agent in the remaining lanes is selected as the middle lane. Using the direction and the closest lane center point to the target agent of the middle lane, a straight line is created to divide the 2D plane into the left region and the right region. A lane will then be selected from each region and, along with the middle lane, the three lanes will be used as the lane input of the model. To be exact, for each remaining lane, say $L$, let the point among its center points closest to the target agent be $l_a$. Then the lane will be associated with the same region where the point $l_a$ is located. After that, all of the remaining lanes (except the middle lane) will be associated with either the left region or the right region. For all lanes in a region, the one with its $l_a$ closest to the straight line is selected to be included in the lane input.
    
    Note that, in some cases, there may be no lane in the left or right region. This will lead to the lane input to contain fewer than 3 lanes. If this situation occurs, the missing lanes (which could be the lane in the left region, the lane in the right region, or both) will be padded in the lane input. After the padding, a lane mask is used to ensure that the predicted trajectory does not appear on the padded lane. Let $M_a = \{m_l,m_m,m_r\}$ be the lane mask of target agent $a$, where $m_l, m_m, m_r$ denote the Boolean values representing the existence of the corresponding input lanes. If the lane in the left (right) region exists, $m_l$ ($m_r$) will be True. Otherwise, it will be False. Note that, according to our design, $m_m$ is always True. If a target agent is not located in any lane, the agent will be filtered out in the beginning.
    
    With Figure~\ref{fig:lane_process_4_extend} as an example, there are two lanes to be selected in the lane input. The middle lane is marked in red and the right lane is marked in orange. There is no lane in the left region. As a consequence, the left lane in the model input is padded. After padding, there will still be 3 lanes in the model input. To avoid the predicted trajectory to appear in the left padded lane, the lane mask $M_a = \{ \textrm{F}, \textrm{T}, \textrm{T}\}$ is used to shield the padded lane. 
    
    \noindent\textbf{Lane extension.}
    The goal of our model is to predict the trajectory of a target agent in the next 6 seconds. In urban environment, the vehicle speed is usually restricted to be lower than \SI{50}{\kilo\meter\per\hour}. As a consequence, the farthest distance a target agent can travel in 6 seconds is approximately 80 meters. In the last step, each input lane is extended into 80 meters for inference in the case that if some of them are not long enough in the labelled dataset. 
    
    Given an input lane, starting from the center point closest to target agent, we verify whether there are 17 more center points ahead of the starting one. (Note that the distance of two consecutive center points is 5 meters.) If there are, the lane is considered to have enough distance for inference and nothing needs to be done. 
    
    If not, say there are $k$ more center points ($k < 17$) and the last center point is denoted as $l_k$, the next virtual point, $l^v_{k+1}$, is calculated by adding 5 meters to $l_k$ along the lane direction $\vec{L_k}$.
    
    Using virtual point $l^v_{k+1}$ and the direction orthogonal to $\vec{L_k}$, a line can be drawn which divides the 2-D plane into the front region and the rear region. The lane with a center point nearest to $l^v_{k+1}$ is selected as the next lane candidate. If the next lane candidate has more than two center points in the rear region, it will be discarded and another lane with a center point nearest to $l^v_{k+1}$ is selected as the candidate. This process is repeated until a next lane candidate having fewer than two center points in the rear region is found. If all next lane candidates are discarded, the target agent will be filtered out from the dataset. After that, starting from the first center point, $17 - k$ center points of the next lane candidate will be added to the lane input. If the next lane candidate does not have $17 - k$ center points, the aforementioned process will be repeated until the input lane is extended to 80 meters. An example is shown in Figure~\ref{fig:lane_process_4_extend}. By adding 5 meters to last center point $l_k$ along the lane direction $\vec{L_k}$, the next virtual point, $l^v_{k+1}$, is obtained. Using $l^v_{k+1}$, the next lane candidate is found, which is used to extend the lane input to 80 meters.
\end{enumerate}
    
\begin{figure}[!ht]
\begin{center}
\includegraphics[width=0.4\textwidth]{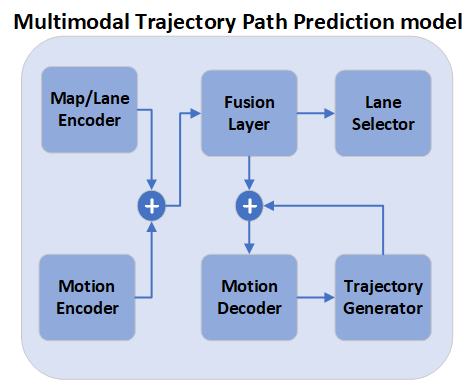}
\caption{Multi-Modal Transformer Path Prediction model structure} 
\label{fig:MTPP_model_structure}
\end{center}
\end{figure}
\subsection{Model architecture}
Figure~\ref{fig:MTPP_model_structure} illustrates our MTPP model architecture. The model is broken down into seven parts, including agent history encoding, map information encoding, features fusion, lane classification, trajectory generation, auto-regression vs non-auto-regression, and loss function. Each will be elaborated as follows.
    \subsubsection{Agent history encoding}
    In agent history encoding, the historical positions of the agent will be passed to the motion encoder. Here, a transformer encoder is adopted with an important modification. First, the embedding layer is replaced with a linear layer because the former is specifically designed for NLP tasks. It acts as a lookup table, which will map each word input to a unique vector representation. However, in our case, the input is already a unique value with a certain meaning (e.g., relative coordinate, speed, acceleration). The transformer encoder has 6 layers, each of which has 8 heads and 512 hidden dimensions for Feed-Forward block.
    \subsubsection{Map information encoding}
    To make use of map information, our system will use a different neural network model depending on the availability of different resolutions of map data. If the map data contain only the occupancy map, 2D CNN will be used as our map encoder, which has 4 layers with filter sizes of {5, 5, 5, 3} and respective strides of {2, 2, 1, 1}.
    If the map data contain the lane feature, 1D CNN will be used as our map encoder, which has 1 layer with filter size 3 and respective stride 2. 
    Both CNNs are followed with a fully connected layer with 32 hidden dimensions.
    \subsubsection{Features fusion}
    After the historical agent positions and map data are encoded, they will be concatenated and then passed to the fusion layer. In the fusion layer, a linear layer is used to linearly transform our encoded inputs into a high dimensional feature. The linear layer has 1 layer with 512 hidden dimensions.
    \subsubsection{Future lane prediction}
    In future lane prediction, a Multiple Layer Perceptron (MLP) and Softmax function are applied in our classification model. The MLP has 2 layers with hidden dimensions of {256, 3}.
    There are two inputs for lane classification, the result of features fusion and lane mask. 
    
    Since some agents will not have 3 drivable lanes, we will use padding method to fix our input vector and create a lane mask. 
    
    To remove unwanted prediction (i.e., predicting to a non-existing lane), an element-wise product between fusion result and lane mask is computed to obtain only the probabilities of existing candidate lanes.
    
    \subsubsection{Trajectory generation}
    Trajectory generation has two parts, motion decoder and trajectory generator, which form a loop structure for regression.
    After the fusion result is obtained, it will be passed to the motion decoder. A Transformer decoder is used as our motion decoder. Similar to the motion encoder, the input preprocessing of the decoder is adjusted exactly as what is done in the motion encoder and it has 6 layers, each of which has 8 heads and 512 hidden dimensions for Feed-Forward block. The Transformer decoder requires two inputs, source and target. The source input is the result of features fusion. The target input is the coordinates relative to the current position and is initialized to be 0. The output of the decoder is passed to the trajectory generator, which is an MLP with 2 hidden layers of dimensions {256, 2}. During forward passing, the result of trajectory generator will be concatenated to the target input for the prediction of the next timestep. 

    \subsubsection{Auto-Regression vs Non-Auto-Regression}
    
\begin{figure}[!ht]
    \begin{center}
    \includegraphics[width=0.4\textwidth]{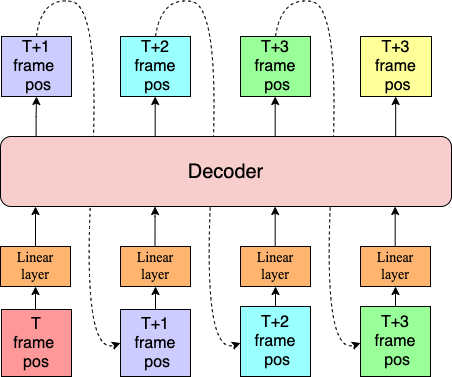}
    \caption{Auto-Regressive Method} 
    \label{fig:auto-regressive}
    \end{center}
    \end{figure}
    
    If a model takes sequential input, it has two ways to train, i.e., auto-regression and non-auto-regression.
    In auto-regression, the output of the prediction result will be concatenated back to input of the model by a loop. An example is provided in Figure~\ref{fig:auto-regressive}, in which $t+1$ predicted frame is copied to the $t+1$ input, and the same can be said for $t+2$ and $t+3$ frames.
    
\begin{figure}[!ht]
    \begin{center}
    \includegraphics[width=0.4\textwidth]{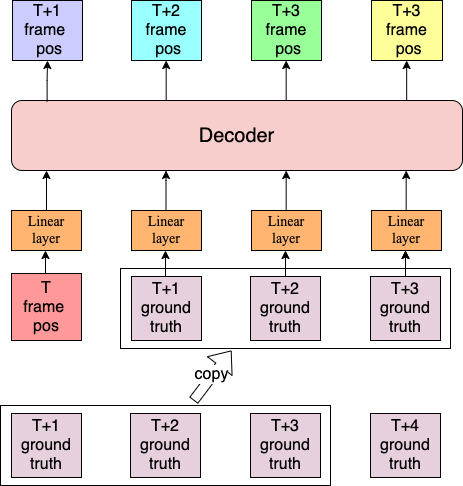}
    \caption{Non-Auto-Regressive Method} 
    \label{fig:non-auto-regressive}
    \end{center}
    \end{figure}
    
    On the contrary, in non-auto-regression, a part of the ground truth will be copied to the input of the model. An example is provided in Figure~\ref{fig:non-auto-regressive} in which $t+1$ to $t+3$ ground truth data are copied to input of the model.
    Non-auto-regression is a popular training method for NLP tasks~\cite{NMT}, because it can reduce the training time to get the inference result without having to go through a loop. Compared to auto-regression, non-auto-regression will rely more on ground truth data to make the predictions. However, auto-regression will consider more past information to predict the result.
     
    \subsubsection{Loss function}
    The loss function is combined by two parts, one is the Mean Square Error (MSE) loss for trajectory prediction; the other is the Cross Entropy (CE) loss for lane prediction. The total loss function is defined as 
    \begin{equation}\label{eq3}
    \centering
    L = \alpha \cdot L_{\mathrm{MSE}}(\hat{y}, y) + (1-\alpha) \cdot L_{\mathrm{CE}}(\hat{l}, l),
    \end{equation}
    where $L_{\mathrm{MSE}}(\hat{y}, y)$ represents the MSE loss for trajectory prediction, $L_{\mathrm{CE}}(\hat{l}, l)$ denotes the CE loss for lane prediction, and $\alpha$ is a weight hyperparameter.

\subsection{Discussion}
    \subsubsection{Auto-regression vs Non-auto-regression}
        In our experiment, the accuracy of auto-regression is surprisingly better than that of non-auto-regression. We think that it is because non-auto-regression will result in the model putting too much focus on target input instead of the history input in the training stage. As a consequence, the accumulated error in the model evaluation affects the performance of the model. Due to this reason, the auto-regression is selected as our training method.
    
    \subsubsection{Issue with Lane Processing}
\begin{figure}[!ht]
        \begin{center}
        \includegraphics[width=0.4\textwidth]{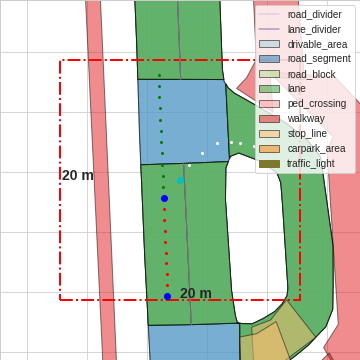}
        \caption{missing label data} 
        \label{fig:fail.png}
        \end{center}
        \end{figure}

    % add scene number 
        In Lane Processing, we discover that some lane information in nuScene is missing. With the 24$^{th}$ training scene in nuScene as an example, Figure~\ref{fig:fail.png} shows its road structure. 
        The points in white are the future trajectory of the target agent and the point in cyan is its current position. The target agent is going to turn right, but the nuScenes labeled data does not have the lane label after the right turn. Therefore, our predictions will be constrained by only one straight forward lane. 
        As we mentioned, this is also the reason why we do not submit our prediction result to the leader board. 
        
\section{Performance}
\label{sec:Performance}
\subsection{Environmental Settings}
For model training and testing, the hardware of our environment includes a server with i7-6700 CPU, 64 GB RAM, and two GTX1080Ti graphic cards. The software configuration includes Ubuntu 18.04, and Torch 1.10.0~\cite{Torch}. Two methods are considered in experiment, Trajectron++  ~\cite{Trajectron++} and MTPP. MTPP has three different approaches: basic transformer, transformer with map data and transformer with lane data.

\subsection{Evaluation Metrics}
Two performance metrics are used, including Average Displacement Error (ADE) and Final Displacement Error (FDE) of single trajectory prediction with different predicted temporal horizon $t = {1,2,3,4,5,6}$. These performance metrics are elaborated as follows.

\begin{enumerate}
\item ADE: The average error of the predicted positions among all frames for all target agents.
\item FDE: The average error of the predicted final position of all target agents.
\end{enumerate}

\subsection{Model Training}
For hyperparameter setting, the batch size is set to 256 and learning rate is initialized to 0.0005 with the decay rate of 0.9999. The nuScene dataset is divided into the training set, validation set and testing set with the ratio of $8 : 1 : 1$. All datasets are processed by lane processing procedure.

\subsection{Experimental Results and Analysis}
In the following sections, the results between MTPP trained with different methods, Auto-Regression (AR) and Non-Auto-Regression (NAR), as well as the results of MTPP and the baseline model (i.e., Trajectron++) are presented and analyzed.

\begin{table}[!ht]
    \centering
    \caption{ADE results of different methods}
    \begin{tabular}{|c|c|c|c|c|c|c|}
    \hline
    \multirow{2}{*}{Method} & \multicolumn{6}{c|}{ADE} \\
    \cline{2-7}
    & $1s$ & $2s$ & $3s$ & $4s$ & $5s$ & $6s$ \\
    \hline
    NAR & $\textbf{0.26}$ & $0.48$ & $0.84$ & $1.38$ & $2.10$ & $3.00$ \\
    \hline
    AR & $0.26$ & $\textbf{0.45}$ & $\textbf{0.69}$ & $\textbf{0.99}$ & $\textbf{1.35}$ & $\textbf{1.77}$ \\
    \hline
    \end{tabular}
    \label{tab:ADE_mtpp_mode}
\end{table}

\begin{table}[!ht]
    \centering
    \caption{FDE results of different methods}
    \begin{tabular}{|c|c|c|c|c|c|c|}
    \hline
    \multirow{2}{*}{Method} & \multicolumn{6}{c|}{FDE} \\
    \cline{2-7}
     & $1s$ & $2s$ & $3s$ & $4s$ & $5s$ & $6s$ \\
    \hline
    NAR & $\textbf{0.32}$ & $0.88$ & $1.92$ & $3.55$ & $5.78$ & $8.57$ \\
    \hline
    AR & $0.34$ & $\textbf{0.76}$ & $\textbf{1.39}$ & $\textbf{2.20}$ & $\textbf{3.22}$ & $\textbf{4.39}$ \\
    \hline
    \end{tabular}
    \label{tab:FDE_mtpp_mode}
\end{table}

\subsubsection{Auto-Regression vs Non-Auto-Regression}
The ADE and FDE of MTPP trained with AR and NAR (both make use of lane information) are shown in Table~\ref{tab:ADE_mtpp_mode} and Table~\ref{tab:FDE_mtpp_mode}, respectively. As can be seen in the tables, MTPP trained with auto-regression significantly outperforms the one trained with non-auto-regression significantly. The possible explanation is that MTPP trained with NAR puts too much focus on target input instead of the history input in the training stage. The accumulated error in the model evaluation affects the performance of the model. Hence, in the following sections, the results of MTPP are all trained with AR.
\begin{table}[t!]
    \centering
    \caption{ADE results of each model}
    \begin{tabular}{|c|c|c|c|c|c|c|}
    \hline
    \multirow{2}{*}{Model} & \multicolumn{6}{c|}{ADE} \\
    \cline{2-7}
     & $1s$ & $2s$ & $3s$ & $4s$ & $5s$ & $6s$ \\
    \hline
    Trajectron++ & $\textbf{0.09}$ & $\textbf{0.34} $ & $\textbf{0.46}$ & $\textbf{0.80} $ & $\textbf{1.24} $ & $1.79 $ \\
    \hline
    MTPP & $0.26$ & $0.45$ & $0.69$ & $0.99$ & $1.35$ & $\textbf{1.77}$ \\
    \hline
    \end{tabular}
    \label{tab:ADE}
\end{table}

\begin{table}[t!]
    \centering
    \caption{FDE results of each model}
    \begin{tabular}{|c|c|c|c|c|c|c|}
    \hline
    \multirow{2}{*}{Model} & \multicolumn{6}{c|}{FDE} \\
    \cline{2-7}
    & $1s$ & $2s$ & $3s$ & $4s$ & $5s$ & $6s$ \\
    \hline
    Trajectron++ & $\textbf{0.09}$ & $\textbf{0.71} $ & $\textbf{1.15}$ & $\textbf{2.14}$ & $3.49 $ & $5.20 $ \\
    \hline
    MTPP & $0.34$ & $0.76$ & $1.39$ & $2.20$ & $\textbf{3.22}$ & $\textbf{4.39}$ \\
    \hline
    \end{tabular}
    \label{tab:FDE}
\end{table}

\subsubsection{MTPP vs Trajectron++}
The ADE and FDE of MTPP with lane information and Trajectron++ are shown in Table~\ref{tab:ADE} and Table~\ref{tab:FDE}, respectively. As can be observed in the tables, MTPP performs better than Trajectron++ in long-term predictions and Trajectron++ has a better performance than MTPP in short-term predictions. There are two possible reasons for the results in the long-term predictions. First, Transformer is better than CVAE at decoding long sequential data. The second reason is the lane information, which better restrains the prediction to possible driving regions. For short-term predictions, the reason why Trajectron++ outperforms MTPP is because of the unicycle model used in Trajectron++.

\begin{figure}[!ht]
\begin{center}
    \includegraphics[width=0.4\textwidth]{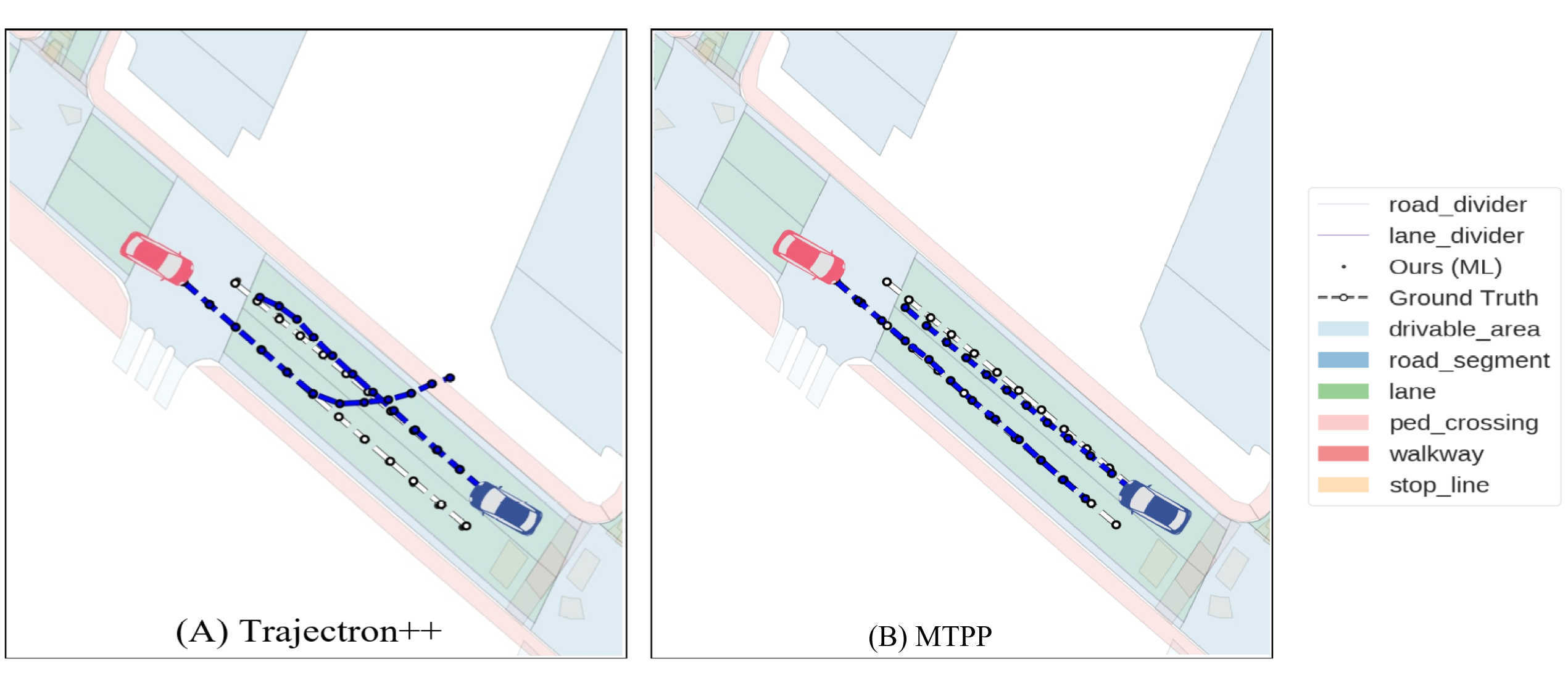}
    \caption{MTPP vs Trajectron++ in a simple scene} 
    \label{fig:simple_result}
\end{center}
\end{figure}

\begin{figure}[!ht]
\begin{center}
    \includegraphics[width=0.4\textwidth]{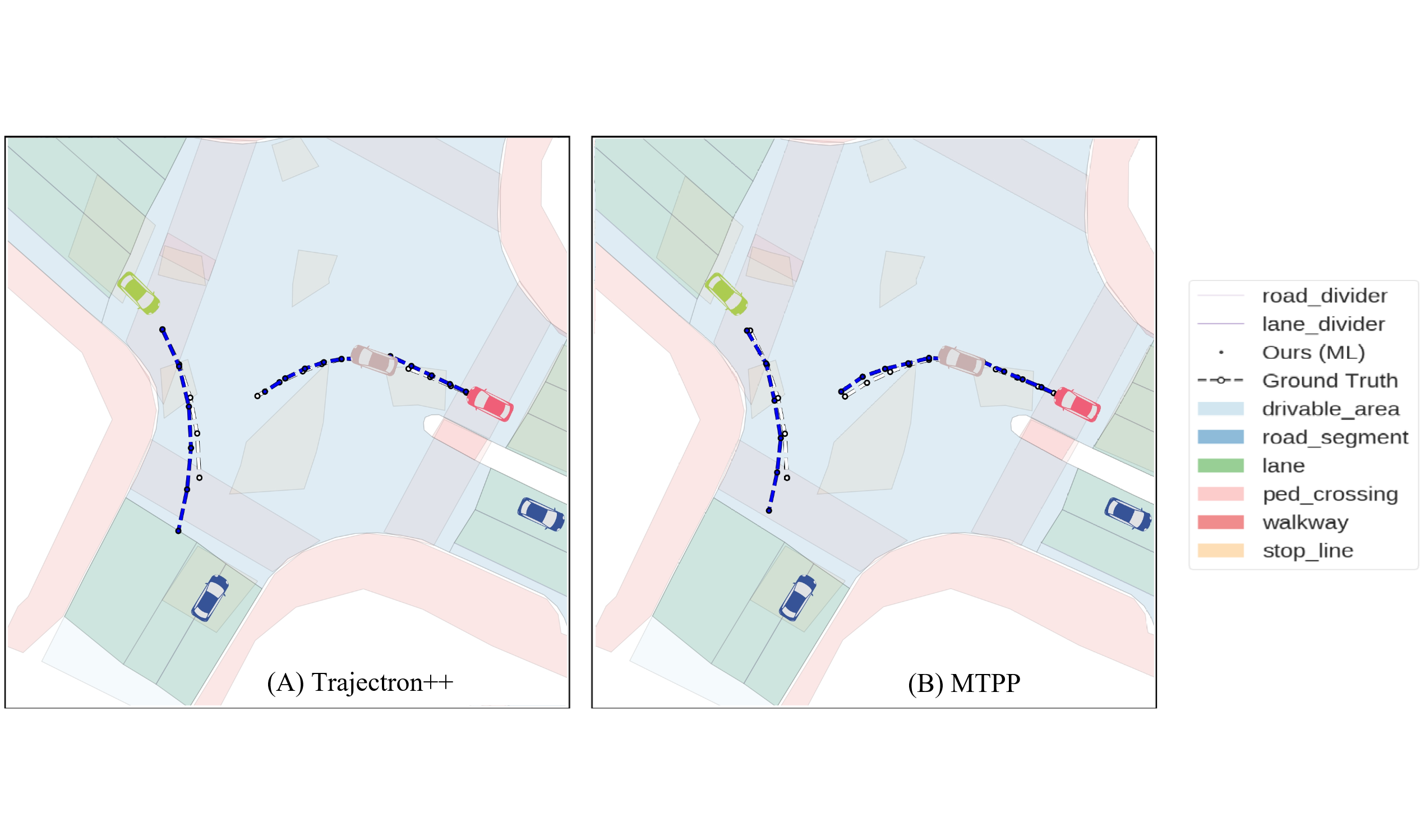}
    \caption{MTPP vs Trajectron++ in a complex scene} 
    \label{fig:complex_result}
\end{center}
\end{figure}

Besides the quantitative results, the visualized results of two scenes are also provided. 
A simple scene is shown in Figure~\ref{fig:simple_result} and a complex scene in Figure~\ref{fig:complex_result}.
As can observed in the simple scene, the predicted trajectory of the red vehicle by Trajectron++ with Map is a right turn, which does not match with the ground truth. On the contrary, the predicted trajectories made by MTPP with Lane input both follow the lane and are more consistent with the ground truth. In the complex scene, both Trajectron++ with Map and MTPP with lane input predict correct turns for vehicles at the intersection and show comparable performance.

\begin{table}[t!]
        \centering
        \caption{FDE results of each model}
        \begin{tabular}{|c|c|c|c|c|c|c|}
        \hline

        \multirow{2}{*}{Model} & \multicolumn{6}{c|}{FDE} \\
        \cline{2-7}
         & $1s$ & $2s$ & $3s$ & $4s$ & $5s$ & $6s$ \\
        \hline
        \small{MTPP (w/o map)} & $\textbf{0.29}$ & $\textbf{0.65}$ & $1.50$ & $2.35$ & $3.55$ & $4.51$ \\
        \hline
        \small{MTPP (w/ map)} & $\textbf{0.29}$ & $0.70$ & $\textbf{1.34}$ & $2.21$ & $3.26$ & $4.46$ \\
        \hline 
        \small{MTPP (w/ lane)} & $0.34$ & $0.76$ & $1.39$ & $\textbf{2.20}$ & $\textbf{3.22}$ & $\textbf{4.39}$ \\
        \hline
        \end{tabular}
        \label{tab:ablation_fde}
\end{table}

\subsection{Ablation Study}
In this section, the impacts of map and lane information to the model prediction are presented. The FDE of MTPP without map, with map, and with lane information are provided in Table~\ref{tab:ablation_fde}, respectively. As can observed in the table, MTPP with map is better than the one without map and MTPP with lane is better than the one with merely map in the long-term predictions. These results are consistent to our expectation. However, the map and lane information does not help much in the short-term predictions. This is because, within a short period of time, the vehicle is unlikely to go out of the boundary of the lane or the road map.

\section{Conclusion}
In this thesis, a path prediction system, MTPP, for autonomous vehicles is proposed. In MTPP, the Transformer architecture is adopted in our model to effectively encode target agent's history input and decode the features. In addition, the lane information is added to the model input, which can confine the prediction result in a more reasonable area in the long run. For data preprocessing, the lanes which are in opposite direction to target agent are deleted. Moreover, lane mask is applied to remove the missing lane in the lane input. To ensure the length of the lane input is long enough, the consecutive lane chunks are carefully stitched together. The proposed system is evaluated by nuScene dataset and is measured by average distance error and final distance error. The result shows that MPTT is better than Trajectron++ in long-term prediction. In ablation study, the additional lane feature helps MPTT to achieve better results.

For the future work, we hope to make our prediction more accurate. To accomplish this goal, more features could be added to our MTPP model, such as the direction signal of the target agent. In addition, other loss functions, such as the lateral distance error between center lane position and prediction result, can be included to better enforce the predicted position fall into the drivable area.

\bibliography{reference}

\end{document}